\documentclass[journal]{IEEEtran}
\usepackage{amsmath,amsfonts}
\usepackage{colorprofiles}
\usepackage{svg}
\usepackage{array}
\newcolumntype{Y}{>{\centering\arraybackslash}X}
\usepackage{xcolor}
\usepackage{makecell} % 加载 makecell 宏包
\usepackage{algorithmic}
\usepackage[ruled,linesnumbered]{algorithm2e}
\usepackage{array}
\usepackage[caption=false,font=footnotesize,labelfont=rm,textfont=rm]{subfig}
\usepackage{textcomp}
\usepackage{stfloats}
\usepackage{url}
\usepackage{verbatim}
\usepackage{graphicx}
\usepackage{tabularx}

\usepackage{color, soul} 
\usepackage{booktabs}
\usepackage{multirow}
\usepackage[pagebackref=false,breaklinks=false,linkcolor=blue,anchorcolor=black,citecolor=blue,citebordercolor=blue,urlbordercolor=blue,colorlinks=true,bookmarks=true,pdfencoding=auto]{hyperref}
\usepackage{cite}
\setulcolor{blue}

\begin{document}

\title{Light-ResKAN: A Parameter-Sharing Lightweight KAN with Gram Polynomials for Efficient\\ SAR Image Recognition}
\author{Pan Yi, Weijie Li, Xiaodong Chen, Jiehua Zhang, Li Liu, and Yongxiang Liu

\thanks{ The authors are with the College of Electronic Science and Technology, National University of Defense Technology, Changsha, 410073, China.\par
Corresponding author: Weijie Li (lwj2150508321@sina.com).
\par 
This work was supported by National Natural Science Foundation of China (NSFC) under Grant 62376283 and 62531026, Innovation Research Foundation of National University of Defense Technology (JS2023-03), and the Fundamental and Interdisciplinary Disciplines Breakthrough Plan of the Ministry of Education of China (JYB2025XDXM110). Emails: Pan Yi  (yipan@nudt.edu.cn), Jiehua Zhang (Jiehua.Zhang@oulu.fi), Xiaodong Chen (cxd@nudt.edu.cn), Li Liu (liuli$ \_ $nudt@nudt.edu.cn) and Yongxiang Liu (lyx$ \_ $bible@sina.com). }
}

\maketitle

\begin{abstract}
Synthetic Aperture Radar (SAR) image recognition is vital for disaster monitoring, military reconnaissance, and ocean observation. However, the large size of SAR images, a result of their unique imaging perspective, hinders the deployment of deep learning models on resource-constrained edge devices. Existing lightweight models often struggle to balance high-precision feature extraction with low computational requirements. The emerging Kolmogorov-Arnold Network (KAN) significantly enhances fitting capabilities by replacing fixed activation functions in traditional neural networks with learnable ones and theoretically requires only 2n+1 activation functions to approximate arbitrary functions, drastically reducing both parameter count and computational overhead. Inspired by KAN, we propose Light-ResKAN to achieve a better balance between high-precision feature extraction and low computational requirements. Firstly, Light-ResKAN fundamentally modifies ResNet by replacing convolutional layers with KAN convolutions, which allows for more intricate and adaptive feature extraction in SAR images. In addition, we leverage Gram Polynomials as activation functions, which are particularly well-suited for SAR data due to their ability to efficiently capture complex non-linear relationships. To further enhance computational efficiency, we innovatively employ a parameter-sharing strategy, where each convolutional kernel shares a set of parameters within the same channel, enabling each channel to learn unique features while significantly reducing the model's parameter count and Floating Point Operations (FLOPs). The model achieved a recognition accuracy of 99.09\%, 93.01\%, and 97.26\% on the MSTAR, FUSAR-Ship, and SAR-ACD datasets, respectively. Experiments on MSTAR resized to 1024×1024 show that compared to the traditional VGG16 model, our model achieves an 82.90-fold reduction in FLOPs and an 163.78-fold reduction in parameter count.  In conclusion, this work establishes an efficient solution for SAR image recognition in edge computing scenarios.

\end{abstract}


\section{Introduction}
\label{Introduction}

\IEEEPARstart{S}{ynthetic} Aperture Radar (SAR) is a radar system that produces high-resolution imagery for Earth observation by transmitting electromagnetic pulses and processing the returned signals\cite{Wiley1985}. Capable of all-weather and all-day operation, SAR actively supports extensive applications in environmental monitoring and defense, spanning hydrology, topography, resource management, and disaster response \cite{ref1,xu2024cloudseg,jiang2025high}.
SAR image recognition is the technology for automated detection and identification of various targets and patterns in SAR imagery through computational methods. In recent years, the rapid advancements in deep learning technology have offered technical avenues for SAR target recognition \cite{Fein-Ashley_2023, li2024predicting, LiSARATRX25} and various deep neural networks have significantly propelled research progress in this domain through their proficient feature extraction capabilities \cite{datcu2023explainable,hua2015deep, 726791, 10533864, hinton2006reducing,11146877,11072100, williams2013gradient, goodfellow2014generative, vaswani2017attention, liu2024s, Liu2025Causal}. 
These methods learn deep feature representations from images, enabling high-accuracy classification and detection tasks \cite{Yan_2024_11, zhou2024madinet,10977792, zhou2024diffdet4sar}.

\par

However, their substantial model complexity and high computational resource demands present significant limitations for deployment in resource-constrained environments, such as unmanned aerial vehicles (UAVs) and satellites \cite{Housseini_2017, zhou2025fiftyyearssaratr}. Concurrently, SAR images captured by UAVs or satellite platforms often exhibit larger spatial extents and pixel dimensions compared to conventional RGB imagery. 
This characteristic, stemming from the need for wide geographical coverage, leads to a rapid escalation in computational workload and resource consumption. 
\emph{Consequently, the development of lightweight models for SAR image recognition has become a paramount research focus.} 
\par
Conventional model compression techniques—including network pruning, knowledge distillation, quantization and low-rank approximation—have been widely explored \cite{Li_2023, Zhou_2025_04, su2023lightweight,yang2024dynamic,Choudhary_2020_02,Li_2023_03}. A dominant trend involves drawing inspiration from lightweight architectures such as MobileNet \cite{howard2019searching} and ShuffleNet \cite{ma2018shufflenet}, which heavily employ depthwise separable convolutions to enhance parameter and computational efficiency. Another line of work adapts mechanisms from vision transformers, such as the hierarchical window attention in Swin Transformer \cite{liu2021swin}, to significantly reduce computational overhead. Therefore, mainstream research has primarily focused on refining key components within established CNN and Transformer architectures. Nevertheless, although they achieved significant results in reducing model size and computational burden, they often come at the cost of sacrificing recognition accuracy due to the loss of high-frequency information caused by model compression, limiting overall performance and robustness \cite{18,21}. \emph{This inherent trade-off presents a persistent challenge in deploying efficient yet highly performant deep learning models in resource-constrained environments.}

The emerging Kolmogorov-Arnold Network (KAN) \cite{7} enhances model expressiveness and interpretability through learnable activation functions, fundamentally adhering to the mathematical principles of Kolmogorov's superposition theorem. Theoretically, this architecture requires only 2n+1 activation functions to exactly approximate arbitrary continuous functions. Compared to conventional deep learning frameworks, KAN demonstrates a substantial parameter efficiency advantage. While maintaining nonlinear approximation capabilities, it is more parameter-efficient than traditional MLPs and circumvents the necessity for stacked convolutions in CNNs and the quadratic growth in computational complexity inherent in Transformer architectures. This architectural innovation has been widely implemented in remote sensing image recognition tasks, demonstrating practical efficacy across diverse application scenarios \cite{240, 25, 26, KAN_ECCV2024}. The ability of KANs to model complex, non-linear relationships efficiently makes them a strong candidate for tackling the intricate patterns present in SAR data for resource-constrained environments \cite{Granata2024}. 
\par

Capitalizing on the advantages of KAN, this paper introduces Light-ResKAN, a novel lightweight architecture for SAR image recognition that integrates ResNet with KAN convolutions. Our approach fundamentally modifies the ResNet architecture by replacing the linear transformations within convolutional layers with learnable, nonlinear activation functions based on KANs. \textbf{To enhance feature representation}, we employ Gram Polynomials as activation functions due to their strong suitability for capturing the complex scattering characteristics of SAR data. By integrating the expressive power of Gram Polynomials within KAN framework, our model demonstrates a more accurate approximation of high-frequency components. Consequently, this approach leads to improved recognition accuracy across multiple SAR datasets. \textbf{To ensure computational efficiency}, we introduce a channel-wise parameter-sharing strategy in the KAN convolutions. In this configuration, each channel of the convolutional kernel shares a unified set of activation function parameters, enabling different channels to learn complementary features. This approach results in substantial reductions in both the number of parameters and FLOPs. 
\par

\begin{figure}[!t]
\centering
\includegraphics[width=0.49\textwidth]{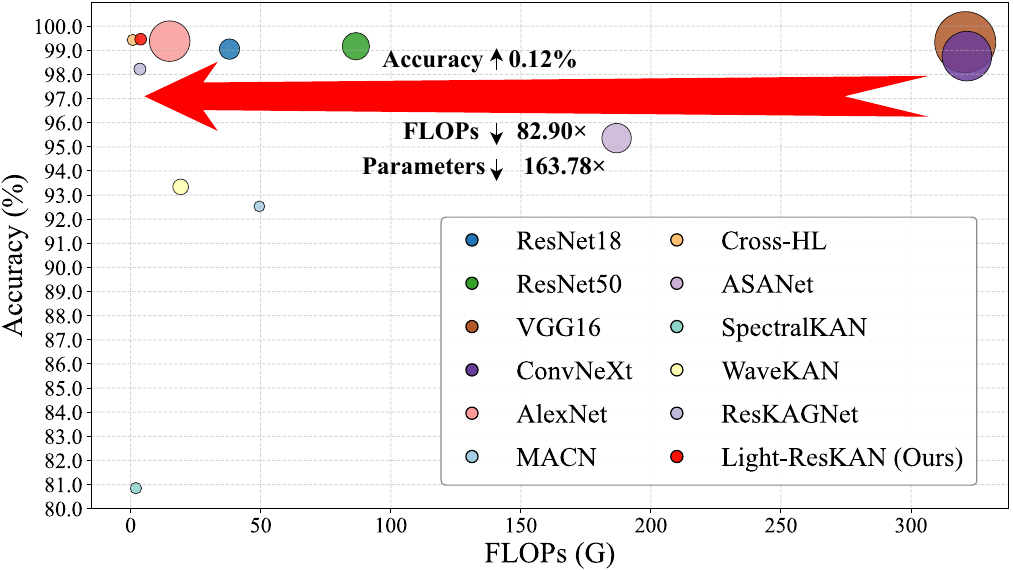}
\caption{
Comparison of our proposed method with existing methods in terms of FLOPs and accuracy. The size of the circle represents the parameter size of the method. The graph shows the excellent balance between accuracy and computational resources of our method evaluated on MSTAR dataset resized to $1024 \times 1024$. Compared to the traditional VGG16 architecture, Light-ResKAN achieves a 0.12\% increase in accuracy while demonstrating significant computational and parameter efficiency, evidenced by an 82.90 times reduction in FLOPs and a 163.78 times reduction in parameter count. }
\label{figure1}
\end{figure}

With a remarkable efficiency of just 0.82 M parameters and a 2.41 MB model size, Light-ResKAN attained accuracies of 99.09\%, 93.01\%, and 97.26\% on the MSTAR, FUSAR-Ship, and SAR-ACD datasets, respectively. Furthermore, with a batch size of 16 and input images of $112 \times 112$, the model's FLOPs per batch were as low as 0.05 G. Besides, as depicted in Fig. \ref{figure1}, our proposed Light-ResKAN model achieves the highest recognition accuracy among all compared models on MSTAR dataset resized to $1024 \times 1024$. Crucially, this superior performance is realized with a significantly reduced model complexity, characterized by a substantially lower parameter count and fewer FLOPs. Specifically, compared to the traditional VGG16 architecture, the accuracy of Light-ResKAN is 0.12\% higher, FLOPs are significantly reduced by 82.90 times, and the number of parameters is even reduced by 163.78 times. Both quantitative metrics and visual analyses confirm that our method effectively balances model complexity and recognition performance, providing an efficient solution for resource-constrained SAR image recognition.

\par Our contribution can be summarized as follows.\par

\begin{enumerate}
\item {We pioneer the integration of KAN into SAR image recognition by proposing Light-ResKAN, a lightweight deep learning architecture that combines residual learning principles with KAN-based convolutional operations. The proposed methodology implements parameter sharing across convolutional channels to achieve architectural compression. To the best of our knowledge, this represents the first lightweight implementation of KAN architecture specifically tailored for SAR imaging applications.}
\item {The powerful feature extraction ability of Gram Polynomials of Light-ResKAN's design is instrumental in yielding exceptional robustness against speckle noise, a pervasive challenge in SAR imaging. Experimental results show that Light-ResKAN maintains higher recognition accuracy than baseline models under noisy conditions, highlighting its ability to effectively differentiate signal from noise.}
\item {The parameter sharing strategy induces a drastic reduction in both the model's parameter counts and FLOPs. By eliminating redundant parameters and computations, this approach significantly alleviates the computational burden and reduces the model footprint, thereby enhancing its suitability for deployment on resource-constrained edge devices.}

\end{enumerate}
\par

The rest of this article is organized as follows. Section \ref{Related Work} provides a review of lightweight models in SAR image recognition and delves into the related work of KANs. Section \ref{Method} details the architectural framework of Light-ResKAN. Section \ref{EXPERIMENTS AND ANALYSIS} presents extensive experimental results and analyses to demonstrate the superiority, advantages, and limitations of our proposed method. Section \ref{ABLATION EXPERIMENTS} are some discussions. Finally, Section \ref{CONCLUSION AND FUTURE WORKS} concludes this article.

\section{Related Work}
\label{Related Work}

\subsection{Lightweight Models in SAR Image Recognition }

\subsubsection{Lightweight CNN-Based Approaches}
A dominant trend involves modifying traditional CNNs to reduce computational costs. Shao et al. \cite{18} replaced standard convolutions with depthwise separable convolutions, significantly improving efficiency in SAR target recognition. Similarly, Liu et al. \cite{21} integrated an improved MobileNetV2 into YOLOv4-LITE, leveraging depthwise separable convolutions and an enhanced Receptive Field Block (RFB) structure to boost detection speed while maintaining accuracy. Anjali et al. \cite{22} further enhanced MobileNetV2 by incorporating a self-attention mechanism, using an ensemble method with a maximum selector for final classification. Xu et al. \cite{23} introduced Mobile-Shuffle, combining hardware-friendly modules and structural reparameterization techniques to accelerate inference.\par
\subsubsection{ Task-Specific Optimization and Multi-Stage Learning}
Some studies focus on optimizing feature learning for specific SAR tasks. Yang et al. \cite{19} proposed R-RetinaNet, which decouples feature optimization for different tasks to mitigate conflicts between learning objectives. Liu et al. \cite{20} adopted a two-stage approach: first, a region-restricted PCA-KMeans clustering algorithm provided pre-classification results, followed by a lightweight MobileNet for refinement, improving overall recognition performance.\par 
\subsubsection{Transformer-Based Lightweight Models}
While most work relies on CNNs, Zhao et al. \cite{zhao24} explored Vision Transformers (ViTs) for SAR processing, proposing a Lightweight Visual Transformer (LViT) that encodes images via a ViT backbone and processes features through an MLP for efficient inference.

Although the above methods have made remarkable progress in improving the efficiency and accuracy of target recognition in SAR images, they each have some limitations. Although the depth separable convolution used in \cite{18} and \cite{20}, \cite{21} and \cite{23} effectively reduce the computational cost, it is not enough to deal with the extraction of high-precision features in a complex electromagnetic environment, and it is easy to lose the unique high-frequency scattering features of SAR images.  The proposed R-RetinaNet 
by \cite{19} mitigates the conflict between learning objectives by decoupling the feature optimization process for different tasks.
However, this approach may increase model complexity and may not be friendly to resource-constrained edge devices. The method of \cite{22} relies on the fine-tuning of the MobilenetV2 architecture of the self-attention mechanism as a binary classifier, which improves the classification accuracy to some extent, but may lack generality and flexibility for multiclass SAR image recognition problems. \cite{zhao24} uses the ViT encoder combined with MLP to realize the lightweight visual transformer (LViT). Although it has advantages in handling long-distance dependencies, it still faces challenges in real-time and computational efficiency.

These specific shortcomings reflect broader, more fundamental limitations of conventional CNNs and Transformers in modeling the complex nonlinear mappings required for SAR imagery. While CNNs excel at capturing local spatial features, their extensive stacking of convolutional layers often leads to prohibitive computational complexity and increased memory burden. Transformers, though effective in global dependency modeling, suffer from the quadratic complexity of their self-attention mechanism, which becomes a significant bottleneck for high-resolution SAR data. Furthermore, both architectures typically require extensive hyperparameter tuning, custom architectural adjustments, and protracted retraining cycles \cite{Xu_2025_07, Rakesh_2025}, which can undermine their generalizability and robustness against SAR-specific challenges like speckle noise and significant resolution variations \cite{Lee2023ResourceEfficient}. This analysis underscores a pressing need for novel, inherently lightweight architectural designs that possess stronger local feature extraction capabilities, superior parameter efficiency, and enhanced nonlinear expressive power beyond the conventional paradigms.

\subsection{Related Work of KAN}
\subsubsection{The Principle of Kolmogorov-Arnold Network}
Kolmogorov-Arnold Network (KAN) is based on the Kolmogorov Arnold theorem, which provides a method for representing a multivariate continuous function through the composition of a finite number of univariate functions and binary addition operations \cite{7}:

\begin{equation}
f(x) = f(x_1, x_2, \ldots, x_n) = \sum_{q=1}^{2n+1} \Phi_q \left( \sum_{p=1}^{n} \varphi_{q,p}(x_p) \right)
\end{equation}
Where $\Phi_q$ and $\varphi_{q,p}$ are single-variable nonlinear functions, n represents the number of input elements, and x denotes the input variable.
\par
KAN enhances model flexibility and expressiveness by replacing fixed linear weights with learnable activation functions on the edges (connections) of the network, while maintaining the interpretability of the model. The activation function in KAN is parameterized by B-spline function, which are piecewise polynomial functions defined by control points and nodes. Each input feature is transformed by a function defined by a spline parameterization, then aggregated into the middle value of each q. Afterwards, it is processed by the function and results in a final output, which is the sum of all these converted values. This approach allows the network to flexibly and efficiently capture intricate data patterns, thereby enhancing adaptability.

\begin{table*}[!t]
    \centering
    \caption{Comparison of CNN, Transformer, and KAN}
    \renewcommand{\arraystretch}{1.5}

    \begin{tabular}{>{\centering\arraybackslash}m{2cm} >{\centering\arraybackslash}m{5cm} >{\centering\arraybackslash}m{4cm} >{\centering\arraybackslash}m{5cm}}
        \toprule
        \makecell{\textbf{Features}} & \makecell{\textbf{CNN}} & \makecell{\textbf{Transformer}} & \makecell{\textbf{KAN}} \\ \midrule
        Computational Complexity & Generally related to image dimensions and kernel size & Quadratic (with respect to sequence length) & Linear (with respect to input dimensionality), dependent on the form of activation functions \\
        Parameter Efficiency & Relatively efficient via weight sharing & Parameters can be large, especially for deep or wide networks & High parameter efficiency depends on the form of the activation functions \\
        Advantages & Strong visual feature extraction, relatively high computational efficiency & Powerful long-range dependency modeling, good parallelism & Low parameter count, computationally efficient, highly interpretable, strong expressive power, easy integration \\
        Disadvantages & Limited global information fusion capability & High computational complexity, large parameter count, challenges in processing long sequences & Emerging field, requires extensive validation and optimization \\ \bottomrule
    \end{tabular}
    \label{tab:comparison_models}
\end{table*}

\par
In KAN, the activation function is a combination of the base function and the spline function SiLU is often used for the basis function, which is defined as $\text{SiLU}(x) = \frac{x}{1 + e^{-x}}$, and the expression for the spline component is $\text{spline}(x) = \sum_i c_i B_i(x)\label{eq:placeholder}$. The B-spline basis function  and coefficients $c_i$ are used here, which are learned during the training process and replace the traditional linear transformation parameters w and b in MLP. Table \ref{tab:comparison_models} provides a comprehensive comparison of the performance and efficiency of different models, including  CNN, Transformer and KAN.

\par

\subsubsection{Application of KAN in Remote Sensing Field}
Minjong Cheon \cite{240} integrated KAN with various pre-trained CNN models to obtain KCN for remote sensing (RS) scene classification tasks. \cite{25} proposed SpectralKAN, which can achieve high hyperspectral image change detection (HSIs-CD) accuracy, while requiring less parameters, FLOPs, GPU memory, training and testing time, and improving the efficiency of the HSIs-CD. \cite{26} proposed a Wavelet-based Kolmogorov-Arnold Network Architecture (WaveKAN) that incorporates Wavelet functions into learnable activation functions for hyperspectral image classification. \cite{KAN_ECCV2024} integrated the KAN layer into the U-Net architecture (U-KAN), segmented farmland using Sentinel-2 and Sentinel-1 satellite images, and analyzed the performance and interpretability of these networks. Table \ref{tab:kan_methods} is a summary of the above methods.\par

\begin{table}[!tb]
    \centering
    \caption{Overview of KAN-Based Methods and Their Application Fields}
    \renewcommand{\arraystretch}{1.5}% Adjusted row height for better readability, adjust as needed
    \setlength{\arrayrulewidth}{0.8pt}% Set line thickness
    % Adjusted column widths for better content distribution. 
    % You might need to fine-tune these values based on your actual document margins and content.
    \begin{tabular}{>{\centering\arraybackslash}m{1.5cm} >{\centering\arraybackslash}m{2.5cm} >{\centering\arraybackslash}m{2cm} >{\centering\arraybackslash}m{1cm}} 
        \toprule
        % Using \makecell for better control over multi-line cells and centering
        \textbf{Methods} & \textbf{Application Fields} & \textbf{Datasets} & \textbf{Activation Functions} \\ \midrule
        KCN & Remote Sensing Scene Classification & EuroSAT & Spline \\ 
        SpectralKAN & Hyperspectral Image Change Detection & Farmland, river, USA; Bay Area, Santa Barbara & Spline \\ 
        WaveKAN & Hyperspectral Image Classification & Salinas, Pavia, Indian Pines & Wavelet \\ 
        U-KAN & Agricultural Land Segmentation & South Africa Crop Type, Sentinel-1 and Sentinel-2 & Spline \\ \bottomrule
    \end{tabular}
    \label{tab:kan_methods}
\end{table}

Despite the significant progress made by SpectralKAN, WaveKAN and U-KAN in their respective fields, these methods still have some limitations, especially when dealing with SAR images. Firstly, the aforementioned studies mainly focus on hyperspectral image classification, remote sensing scene classification or specific types of satellite image segmentation tasks, without fully considering the strong scattering noise and geometric deformation sensitivity issues inherent to SAR images, which limits their direct applicability in the SAR domain. Secondly, although WaveKAN improves model performance by introducing wavelet functions as activation functions, there is insufficient exploration of how to balance the relationship between computational efficiency and model accuracy. Thirdly, regarding dataset selection, many existing works rely on specific types of remote sensing datasets for experimental verification, while using fewer datasets that contain SAR images under complex environments. This means that the generalization ability and adaptability of current models may be limited, making it difficult to cope with challenges in diverse real-world scenarios. 
\par
In view of the above limitations, we propose a new lightweight network framework: Light-ResKAN, which combines Gram Polynomials and KAN convolution. The framework not only enhances the ability to capture complex scattering characteristics in SAR images by embedding KAN's dynamic activation function, but also achieves significant model compression through strategy for sharing parameters within channels, ensuring that the number of parameters is greatly reduced while maintaining high classification accuracy. More importantly, our model has been rigorously tested on various types of SAR image datasets, proving its strong adaptability and efficiency in different application scenarios. Therefore, compared with the existing solutions, Light-ResKAN provides a more flexible and efficient alternative for SAR image recognition.

\section{Methodology}
\label{Method}
Building upon Bottleneck ResKAGNet \cite{8}, Light-ResKAN represents a novel network architecture developed through a series of targeted modifications and optimizations. To overcome the inherent limitations of Bottleneck ResKAGNet, this work elucidates the design motivation and architectural structure of the lightweight model, Light-ResKAN, specifically engineered for the unique properties of SAR images. The section delves into the critical components of Light-ResKAN from diverse viewpoints, encompassing theoretical underpinnings, detailed implementation procedures, and their respective contributions.
\subsection{The Design Motivation of Light-ResKAN}
SAR images exhibit distinct characteristics compared to optical imagery, owing to its unique imaging mechanism and data properties \cite{Guo_2024}. Specifically, SAR images are severely afflicted by coherent speckle noise during the imaging process. Furthermore, due to the influence of side-looking imaging geometric distortions, their spatial resolution is typically lower, commonly ranging between 0.1 and 10 meters \cite{Luo_2023}, \cite{Zhang_2023_01}. This resolution is notably insufficient when contrasted with optical imagery acquired from the same platform.\par

Although the conventional Bottleneck ResKAGNet architecture markedly elevates nonlinear representational capacity through the assignment of discrete Gram Polynomial basis weights to each constituent element of its convolutional kernels, this strategy of element-wise independent parameterization incurs substantial limitations. Specifically, this approach leads to a substantial increase in network parameters and computational load, making it computationally intensive. When processing SAR images with high noise levels, an excessive number of parameters not only escalates model training time considerably but also poses a risk of overfitting to the noise components within the training data, consequently diminishing the model's generalization capability on unseen data. This phenomenon can severely impede the practical deployment and efficiency of SAR image processing systems.\par

To address the aforementioned issues, we propose a novel network architecture: Light-ResKAN. Within the Light-ResKAN architecture, for each convolutional kernel, each channel shares an identical set of weight parameters, thereby enabling each channel of the convolutional kernel to share a learnable activation function. This design offers the following significant advantages:\par

\subsubsection{Parameter Reduction}
By virtue of each channel sharing an identical set of weight parameters, the number of parameters within each convolutional kernel is substantially reduced, thereby decreasing the model's complexity.

\subsubsection{Computational Efficiency}
Light-ResKAN adopts a structure analogous to depthwise separable convolutions to achieve parameter sharing and a significant reduction in FLOPs. This contributes to a more computationally efficient model, particularly for large-scale SAR datasets.

\subsubsection{Mitigation of Overfitting}
Due to the inherent properties of SAR imaging, such as significant speckle noise and potential geometric distortions, traditional models with a high parameter count are prone to overfitting to these imperfections. The Light-ResKAN architecture, by adopting a parameter-efficient design with shared weights within channels, exhibits inherent robustness against such artifacts. This weight sharing mechanism effectively regularizes the model, preventing it from excessively learning noisy or distorted features in SAR data, thereby improving its reliability.

\subsection{The Overall Framework of Light-ResKAN}
\begin{figure*}[!t]
\centering
\includegraphics[width=\linewidth]{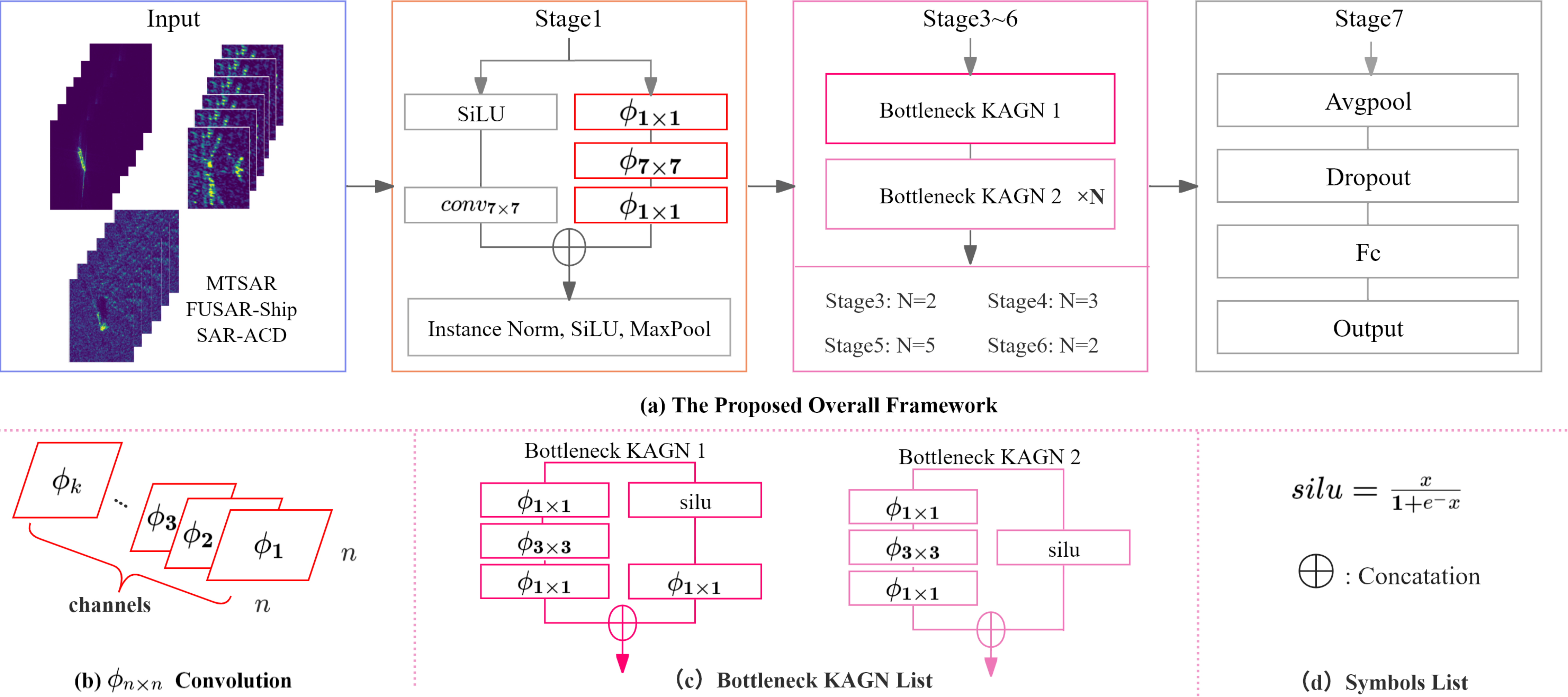}
\caption{
Model architecture diagram of Light-ResKAN. (a) Provides its framework with three main modules: the shared activation function convolution $\phi_{n \times n}$, Bottleneck ResKAGN 1 and Bottleneck ResKAGN 2. (b) Details the internal structure of the shared activation function convolution $\phi_{n \times n}$. (c) Illustrates the architecture of Bottleneck ResKAGN 1 and Bottleneck ResKAGN 2. (d) Provides a symbol List.
}
\label{figure5}
\end{figure*}

The specific design workflow of Light-ResKAN is as follows. Input image data is first processed through separate ordinary convolutional layers and KAN convolutional layers. The outputs from these two types of layers are then linearly combined to fuse features, generating feature maps that encapsulate multi-scale information. Subsequently, these multi-scale feature maps undergo normalization and activation functions to enhance their representational capacity. Following this, a max-pooling operation is applied for downsampling, reducing the spatial dimensions of the feature maps while retaining the most salient feature information. After downsampling, the feature maps are sequentially processed by four Bottleneck ResKAGN blocks. Each of these blocks comprises 3, 4, 6, and 3 Bottleneck ResKAGN convolutional units, respectively. Within these blocks, the first unit is specifically designated as Bottleneck ResKAGN 1, while the remaining units are uniformly labeled as Bottleneck ResKAGN 2. This hierarchical processing through the four Bottleneck ResKAGN blocks further refines and strengthens the information contained within the feature maps. Finally, the processed feature maps are subjected to an average pooling layer for global feature extraction. The resulting features are then fed into fully connected layers for the ultimate classification or regression task. Prior to outputting the final results, dropout technology is applied. By randomly discarding a portion of neurons, this technique enhances the model's generalization capability, leading to more robust predictions. Through this series of meticulously designed steps, Light-ResKAN can efficiently process image data and produce stable and reliable prediction outcomes. Fig. \ref{figure5} shows the Light-ResKAN model architecture.

After presenting the overall architecture of Light-ResKAN, we further elaborate on the two key designs that enable its lightweight nature and strong representational capability. First, we introduce the shared activation convolutions (Gram polynomial-based KAN convolutions) as the fundamental operator. This operator replaces fixed activation functions with learnable polynomial activations, thereby enhancing the nonlinear representation capacity of feature mappings while reducing the risk of overfitting. Building upon this foundation, we develop the bottleneck ResKAGN module, which integrates the shared-activation convolution into a residual bottleneck framework and effectively mitigates the parameter explosion caused by polynomial expansion through a channel-wise parameter sharing mechanism. Consequently, Light-ResKAN achieves sufficient nonlinear modeling capacity for complex SAR target representation while maintaining a compact model footprint suitable for resource-constrained deployment.

\subsection{Implementation Principles of Different Modules}

Light-ResKAN primarily comprises three essential components:

\subsubsection{Shared Activation Function Convolutions}

\begin{figure}[!t]
\centering
\includegraphics[width=\linewidth]{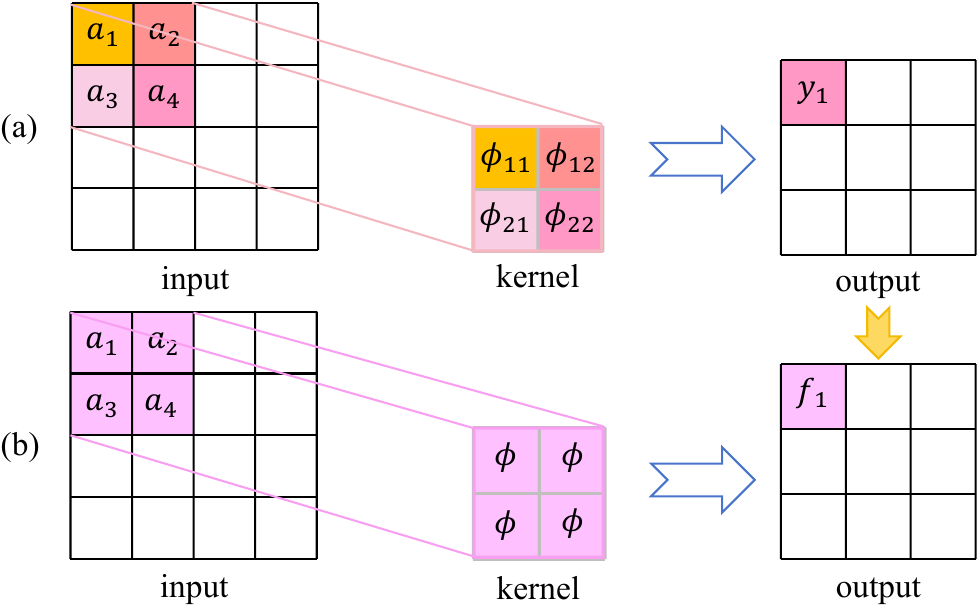}
\caption{
An example of $\phi_{2 \times 2}$ convolution in one channel. (a) The operation process and result of initial KAN convolution. (b) The operation process and result of the proposed $\phi_{2 \times 2}$ convolution with shared weights. Among them, \( {y}_{1} = {\phi }_{11}\left( {a}_{1}\right)  + {\phi }_{12}\left( {a}_{2}\right)  + {\phi }_{21}\left( {a}_{3}\right)  + {\phi }_{22}\left( {a}_{4}\right) \) and \( {f}_{1} = \phi \left( {a}_{1}\right)  + \phi \left( {a}_{2}\right)  + \phi \left( {a}_{3}\right)  + \phi \left( {a}_{4}\right) \)
}
\label{gongshi}
\end{figure}

The initial KAN convolution exhibits parameter redundancy. To address this, our proposed $\phi_{n \times n}$ convolution is designed to reduce the number of parameters and FLOPs. Specifically, 
$\phi_{n \times n}$ achieves parameter sharing within the same channel, where a single set of weights is applied across all spatial locations of a given channel. This effectively treats each channel as a single activation function. Fig. \ref{gongshi} takes a $2 \times 2$ convolution as an example to show the operation process of initial KAN convolution and $\phi_{n \times n}$ convolution on the image in a single channel. This formulation allows a single set of learnable parameters weights to govern the activation behavior across an entire channel, significantly reducing the overall parameter count compared to element-wise learnable kernels.\par

Besides, in the realm of neural networks, conventional convolutional layers typically employ fixed activation functions such as ReLU, Sigmoid, or Tanh. To introduce learnable non-linearities directly within the convolutional kernel itself, the KAN convolution has been proposed. In KAN convolution, each element $\phi$ of the convolutional kernel is not a fixed value, but rather a learnable non-linear function. This function is typically constructed based on a spline representation, with its general form given by:
\begin{equation}
    \phi(x) = \varphi_{\text{spline}}(x) + w_m \cdot b(x)
\label{eq:kan_general}
\end{equation}
where $\varphi_{\text{spline}}(x) = \sum_i c_i B_i(x)$ is the spline component with B-spline basis functions $B_i(x)$ and learnable coefficients $c_i$, and $b(x) = \text{SiLU}(x) = \frac{x}{1+e^{-x}}$ is the residual base function. The kernel slides over the 2D input data, and the function $\phi$ computes each pixel of each image in turn, and then summarizes the results into a single output pixel.
\par

To effectively address the complex scattering characteristics inherent in SAR imagery, and drawing inspiration from the work in \cite{8}, we replaced the spline-based $\varphi_{\text{spline}}(x)$ in the original KAN formulation with Gram Polynomials. This substitution was specifically motivated by the suitability of Gram Polynomials for capturing the intricate patterns found in SAR data, thereby enhancing recognition capabilities.
Specifically, we denote the Gram polynomial-based activation function as $\Phi_G(x)$ to distinguish it from the general KAN activation $\phi(x)$ in Eq.~(\ref{eq:kan_general}). The activation function for each channel is formulated as:
\begin{equation}\Phi_G(x)=\sum_{k=0}^{D}w_k G_k(\tanh(x))+w_m \cdot \text{SiLU}(x)\label{eq:gram_activation}\end{equation}
where $D$ is the polynomial degree, $\{G_k(\cdot)\}_{k=0}^{D}$ are the Gram polynomial basis functions, $\{w_k\}$ are learnable convolution weight parameters, and $w_m$ scales the residual path.

\textbf{Gram polynomial basis functions.} The Gram polynomials are constructed via a three-term recurrence relation. Given an input $\tilde{x} = \tanh(x)$ (normalized to $[-1, 1]$), the polynomial basis functions $\{G_k(\tilde{x})\}_{k=0}^{D}$ are defined as:
\begin{equation}G_0(\tilde{x}) = 1, \quad G_1(\tilde{x}) = \tilde{x}\end{equation}
\begin{equation}G_k(\tilde{x}) = \tilde{x} \cdot G_{k-1}(\tilde{x}) - \beta_k \cdot G_{k-2}(\tilde{x}), \quad k \geq 2\end{equation}
where $\beta_k$ is a learnable recurrence coefficient, enabling the network to adaptively adjust the orthogonal basis structure during training. This design differs from classical Gram-Schmidt orthogonalization where the coefficients are fixed, providing additional flexibility for learning task-specific basis representations.

\textbf{Polynomial degree selection.} The polynomial degree is set to $D = 3$ in all experiments. This generates 4 basis functions ($G_0$ through $G_3$) per input channel, providing sufficient nonlinear expressiveness while keeping the polynomial expansion moderate. Higher degrees (e.g., $D \geq 4$) may introduce numerical instability and overfitting risks due to the Runge phenomenon, especially on the relatively small-scale SAR datasets. Lower degrees (e.g., $D = 2$) would limit the function approximation capability.

\textbf{Weight initialization strategy.} All standard convolution weights and polynomial weights are initialized using Kaiming Uniform initialization (linear mode), ensuring proper gradient flow at initialization. The recurrence coefficients ($\beta_k$) are initialized from $\mathcal{N}(0, \sigma^2)$ with $\sigma = \frac{1}{k^d \cdot C_{\text{in}} \cdot (D+1)}$, where $k$ is the kernel size, $d$ is the spatial dimension, and $C_{\text{in}}$ is the number of input channels. This small-variance initialization ensures that the polynomial bases approximate standard orthogonal polynomials at the beginning of training, providing a stable starting point, while the learnable coefficients gradually adapt to the data distribution during optimization.

\par
As an orthogonal polynomial sequence, Gram Polynomials exhibit theoretical properties that align exceptionally well with the unique challenges of SAR imagery, positioning them as a highly promising activation function for SAR image recognition. Their principal advantage lies in their ability to adaptively integrate the processes of feature extraction and speckle noise suppression through adjustable polynomial degrees—effectively mitigating coherent speckle in homogeneous regions while enhancing critical features at edges and strong scatterers. Meanwhile, they adeptly handle the high dynamic range inherent in SAR data with their oscillatory characteristics. The orthogonality of their basis also contributes to numerical stability and generalization capability. More importantly, as demonstrated by the pioneering work of Liu et al. \cite{Liu2016MRELBP} in texture classification, polynomial like features can effectively capture the texture structure information of images. In addition, research on successful texture classification from random features \cite{liuRandomFeatures} also indicates that methods based on function approximation can demonstrate significant strong potential in texture representation tasks. Consequently, from a theoretical perspective, Gram Polynomials offer a more powerful and intelligent nonlinear transformation compared to fixed functions like ReLU, providing a novel solution for SAR image recognition that combines mathematical rigor with enhanced practical adaptability.
\par

\subsubsection{Bottleneck ResKAGN 1}
Bottleneck ResKAGN 1 is a convolutional unit that comprises two parallel convolutional branches. The first branch consists of a sequence of $\phi_{1 \times 1}$, $\phi_{3 \times 3}$, and $\phi_{1 \times 1}$ convolutions. The second branch involves a SiLU activation followed by a $\phi_{1 \times 1}$ convolution. The final output is obtained by the linear summation of the results from these two branches. Assuming the input has $C_{in}$ channels, the intermediate Bottleneck layer has $C_{b}$ channels, and the output has $C_{out}$ channels. In the first branch, the initial $\phi_{1 \times 1}$ convolution compresses the number of input feature map channels from $C_{in}$ to $C_{b}$, significantly reducing the computational cost of the subsequent $3 \times 3$ convolution. The $\phi_{3 \times 3}$ convolution is primarily responsible for feature extraction. The final $\phi_{1 \times 1}$ convolution expands the channel dimension from $C_{b}$ to $C_{out}$. The second branch serves a role analogous to a residual connection, helping to alleviate gradient vanishing in the network. Within this second branch, the $\phi_{1 \times 1}$ convolution handles potential mismatches in channel dimensions or spatial dimensions between the input and output, ensuring smooth information flow and promoting effective learning.

\subsubsection{Bottleneck ResKAGN 2}
The design of Bottleneck ResKAGN 2 and Bottleneck ResKAGN 1 is characterized by their identical functionalities and highly similar structural composition. The sole distinction lies in Bottleneck ResKAGN 2's omission of the $\phi_{1 \times 1}$ convolution in its second branch. This design choice is motivated by the common practice of spatial downsampling, coupled with a doubling of channel dimensions, typically occurring in the initial Bottleneck ResKAGN block of each new stage. To address the resulting dimensionality mismatch introduced by these transformations, a $\phi_{1 \times 1}$ convolutional layer is specifically incorporated into the skip connection. This layer ensures dimensional consistency. Conversely, subsequent Bottleneck ResKAGN blocks within the same stage, having already achieved consistent dimensionality, no longer require additional dimensional adjustments. Therefore, their skip connections typically employ an identity mapping without convolution, thereby simplifying the architecture and enhancing computational efficiency. This design approach effectively preserves the network's overall performance while optimizing computational resource utilization.
\par

By effectively integrating and coordinating these three core elements, Light-ResKAN delivers both superior model performance and substantial reductions in FLOPs and parameters. This efficient design empowers Light-ResKAN to maintain its task execution capabilities in resource-limited settings, underscoring its unique merits and practical utility.

\section{Experiments and Analyses}
\label{EXPERIMENTS AND ANALYSIS}
This section introduces the three publicly available SAR datasets employed in our experiments. Subsequently, we conduct a comprehensive comparative evaluation, both quantitatively and qualitatively, of the proposed Light-ResKAN against nine existing methods. This analysis aims to thoroughly assess the performance of Light-ResKAN.

\subsection{Dataset Introduction}
The SAR image dataset used in this paper is the MSTAR dataset \cite{30}, FUSAR-Ship dataset \cite{31} and SAR-ACD dataset \cite{32}. The details of these three datasets are introduced below.

\subsubsection{MSTAR Dataset}
MSTAR (The Moving and stationarytarget Acquisition and Recognition) is a dataset carefully collected and publicly released by Sandia National Laboratories in The 1990s \cite{30}. The dataset includes 10 types of ground military targets, including infantry vehicles (BMP2), patrol vehicles (BRDM2), troop carriers (BTR60, BTR70), tanks (T62, T72), howitzers (2S1), bulldozers (D7), trucks (ZIL131) and air defense (ZSU234), including diversified scenes and observation conditions, such as the rotation angle of the target and the ground wiping angle.

The comprehensive dataset comprises two defined partitions: standard operating conditions (SOC) and extended operating conditions (EOC). Within the SOC partition, the Test set is constructed to share identical serial numbers and target configurations with the training set. This controlled setup allows for the introduction of variations solely in the direction and depression angle. The EOC partition, conversely, is engineered to reflect a more challenging and authentic battlefield environment, exhibiting marked discrepancies in the target, ambient environment, and imaging modalities. This experiment tests the model performance under standard operating conditions (SOC).

\subsubsection{FUSAR-Ship Dataset}
The FUSAR-Ship dataset was produced by Fudan University based on SAR images of the ultra-fine strip pattern of the Gaofen-3 satellite. It contains two polarization modes, HH and VV, and contains 15 main ship categories, 98 subcategories and various non-ship targets \cite{31}. The data slices are taken from 126 original Gaofen-3 remote sensing images, covering various sea, land, Coast, river and island scenes. This paper tests the effectiveness of the proposed model using 10 classes of samples.

\subsubsection{SAR-ACD Dataset}
The SAR-ACD dataset was developed by AICyberTeam and has a total of 4322 aircraft targets, covering 6 types of civil aircraft such as A220, A330, A320321, ARJ21, Boeing737, Boeing787 and 14 other aircraft categories. It is a dataset dedicated to aircraft target detection \cite{32}. This experiment uses the open source Class 6 civil aircraft type for performance testing.
\par
\begin{figure*}[!t]
\centering
\includegraphics[width=\textwidth]{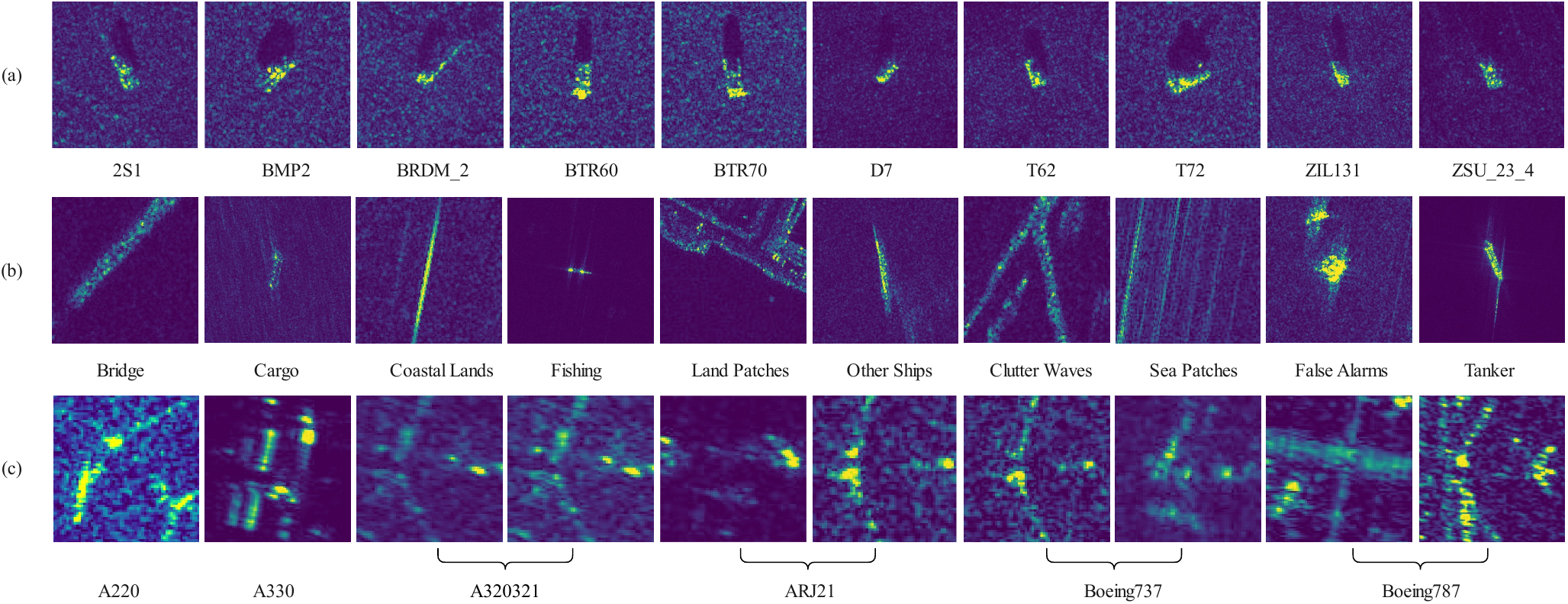}
\caption{Three SAR datasets. (a) Ten types of SAR images in MSTAR. (b) Ten types of SAR images in FUSAR-Ship. (c) Six types of SAR images in SAR-ACD.}
\label{figure6}

\end{figure*}

Table \ref{table2} lists a summary of the MSTAR experimental setup, showing that SAR images with depression angles of 17 ° and 15 ° belong to the training and test sets. Table \ref{tab:fusar_dataset} and Table \ref{aircraft_types} summarizes the experimental setup of FUSAR-Ship and SAR-ACD. Fig. \ref{figure6} shows 10 types of MSTAR dataset, 10 types of FUSAR-Ship dataset, and 6 types of SAR-ACD dataset.

\begin{table}[!t]
    \centering
    \caption{The number of training and testing sets under MSTAR}
    \label{table2}
    % 使用0.5\textwidth来适应半栏宽度
    % \resizebox{\linewidth}{!}{
    \renewcommand{\arraystretch}{1.5} % 调整行高
    % \small
    \begin{tabularx}{\linewidth}{@{}Y|Y|Y|Y @{}}
        \hline
        \multirow{2}{*}{\textbf{Super-classes}} & \multirow{2}{*}{\textbf{Fine-classes}} & \multicolumn{2}{c}{\textbf{Number}} \\ \cline{3-4} 
        & & \textbf{Train (17°)} & \textbf{Test (15°)} \\ \hline
        Rocker launcher & 2S1 & 299 & 274 \\ \hline
        Air defense unit & ZSU23/4 & 299 & 274 \\ \hline
        Trunk & ZIL131 & 299 & 274 \\ \hline
        Bulldozer & D7 & 299 & 274 \\ \hline
        \multirow{2}{*}{\normalfont\textmd{Tank}} & T62 & 299 & 273 \\ \cline{2-4} 
        & T72 & 232 & 196 \\ \hline
        \multirow{4}{*}{\normalfont\textmd{Armored carrier}} & BMP2 & 233 & 196 \\ \cline{2-4} 
        & BRDM2 & 298 & 274 \\ \cline{2-4} 
        & BTR60 & 256 & 195 \\ \cline{2-4} 
        & BTR70 & 233 & 196 \\ \hline
    \end{tabularx}
% }
\end{table}

\begin{table}[!t]
    \centering
    \caption{The number of training and testing sets under FUSAR-Ship}
    \label{tab:fusar_dataset}
    % \resizebox{\linewidth}{!}{
    \renewcommand{\arraystretch}{1.5} % 调整行高
    % \small
    \begin{tabularx}{\linewidth}{@{}Y|Y|Y|Y @{}}
        \hline
        \multirow{2}{*}{\textbf{Super-classes}} & \multirow{2}{*}{\textbf{Fine-classes}} & \multicolumn{2}{c}{\textbf{Number}} \\ \cline{3-4} 
        & & \textbf{Train} & \textbf{Test} \\ \hline
        \multirow{4}{*}{Ships} & Cargo & 366 & 156 \\ \cline{2-4} 
        & Fishing & 248 & 106 \\ \cline{2-4} 
        & Tanker & 150 & 64 \\ \cline{2-4} 
        & Other ships & 312 & 133 \\ \hline
        \multirow{3}{*}{Lands} & Bridges & 1023 & 438 \\ \cline{2-4} 
        & Coastal lands & 707 & 303 \\ \cline{2-4} 
        & Land patches & 1137 & 487 \\ \hline
        Sea & Sea patches & 1250 & 535 \\ \hline
        \multirow{3.5}{*}{Sea scatterer} & Sea clutter waves & \multirow{1.75}{*}{1378} & \multirow{1.75}{*}{590} \\ \cline{2-4} 
        & Strong false alarms & \multirow{1.75}{*}{299} & \multirow{1.75}{*}{128} \\ \hline
    \end{tabularx}
% }
\end{table}

\begin{table}[!t]
    \centering
    \caption{The number of training and testing sets under SAR-ACD}
    \label{aircraft_types} % 添加标签（可选）
    % \resizebox{\linewidth}{!}{
    % \small
    \renewcommand{\arraystretch}{1.5} % 调整行高（只定义一次）
    \begin{tabularx}{\linewidth}{@{}Y|Y|Y @{}}
        \hline
        \multirow{2}{*}{\textbf{Aircraft types}} & \multicolumn{2}{c}{\textbf{Number}} \\ \cline{2-3} 
        & \textbf{Train} & \textbf{Test} \\ \hline
        A220 & 325 & 139 \\
        \hline
        A330 & 359 & 153 \\
        \hline
        A320321 & 357 & 153 \\
        \hline
        ARJ21 & 360 & 154 \\
        \hline
        Boeing737 & 370 & 158 \\
        \hline
        Boeing787 & 353 & 151 \\
        \hline
    \end{tabularx}
    % }
\end{table}

\subsection{Experimental Setup}
\label{setup}
Due to the different initial pixels and types of the three datasets, we take different data processing methods. For FUSAR-Ship, since the image size range is large (88-512), We uniformly resize the image to 512 × 512. For the SAR-ACD dataset, we first resize the image to 128 × 128, and then crop the center of the image to 112 × 112. For fine-grained images such as MSTAR, considering the impact of resize on the recognition target, we cancel resizing and directly crop the image center to 112 × 112. The device used in this experiment is NVIDIA RTX 5090, and the AdamW optimizer is used for gradient descent. The main hyperparameters are set as follows: learning rate = $5\times10^{-4}$, dropout rate = 0.1, batchsize = 16. Each model was trained for 200 full epochs on each dataset using identical hyperparameter settings. All experiments were independently conducted on the three datasets without any dataset-specific tuning, ensuring a rigorous and fair comparison.

%第六张图片

\begin{table*}[!t]
\centering
\caption{Comparison of the Proposed Model with Different Models on Three SAR Datasets.}
\label{tab:comparison}
\renewcommand{\arraystretch}{1.5}
\begin{tabularx}{\linewidth}{@{}c>{\centering\arraybackslash}p{0.5cm} >{\centering\arraybackslash}X >{\centering\arraybackslash}p{1cm} >{\centering\arraybackslash}X *{10}{>{\centering\arraybackslash}X} @{}}
\toprule
\multirow{2.5}{*}{\makecell{Models}} & \multirow{2.5}{*}{\makecell{Year}} & \multirow{2.5}{*}{\makecell{Pars. (M)}} & \multirow{2.5}{*}{\makecell{Size (MB)}} & \multicolumn{3}{c}{Accuracy (\%)} & \multicolumn{3}{c}{FLOPs (G)} & \multicolumn{3}{c}{Test time (ms)} \\
\cmidrule{5-13}
& & & & \multirow{1.5}{*}{\makecell{MSTAR}} & FUSAR-Ship & SAR-ACD & \multirow{1.5}{*}{\makecell{MSTAR}} & FUSAR-Ship & SAR-ACD & \multirow{1.5}{*}{\makecell{MSTAR}} & FUSAR-Ship & SAR-ACD \\
\midrule
AlexNet \cite{alexnet} & 2012 & 57.04 & 228.18 & 97.44 & 89.41 & 95.76 & 0.19 & 3.71 & 0.19 & \textbf{9.54} & \textbf{30.32} & \textbf{9.53} \\
VGG16 \cite{vgg} & 2015 & 134.30 & 537.21 & 96.59 & 90.73 & 96.17 & 4.07 & 80.37 & 4.07 & 11.42 & 43.72 & 11.40 \\
ResNet18 \cite{resnet} & 2016 & 11.18 & 44.73 & 96.71 & 83.54 & 95.34 & 0.49 & 9.51 & 0.49 & 10.31 & 32.41 & 10.29 \\
ResNet50 \cite{resnet} & 2016 & 23.53 & 94.11 & 97.13 & 86.71 & 96.69 & 1.08 & 21.35 & 1.08 & 10.92 & 32.93 & 10.91 \\
ConvNeXt \cite{convnext} & 2022 & 87.58 & 350.23 & 96.04 & 77.86 & 95.83 & 4.07 & 80.53 & 4.07 & 17.23 & 53.25 & 17.19 \\
MACN \cite{macn} & 2022 & \textbf{0.05} & \textbf{0.16} & 82.46 & 89.15 & 79.30 & 0.63 & 12.40 & 0.63 & 97.46 & 296.75 & 97.46 \\
Cross-HL \cite{cross-hl} & 2024 & 0.52 & 2.06 & 98.19 & 78.06 & 97.03 & \textbf{0.01} & \textbf{0.22} & \textbf{0.01} & 11.57 & 35.64 & 11.56 \\
ASANet \cite{asanet} & 2024 & 27.95 & 111.28 & 91.65 & 78.84 & 96.81 & 2.37 & 46.82 & 2.37 & 18.42 & 47.23 & 18.41 \\

SpectralKAN \cite{25} & 2024 & 0.51 & 2.02 & 80.85 & 81.21 & 75.55 & 0.02 & 0.41 & 0.02 & 10.62 & 48.02 & 10.61 \\

WaveKAN \cite{26} & 2024 & 4.82 & 19.23 & 93.34 & 70.61 & 92.51 & 0.02 & 0.40 & 0.02 & 11.41 & 57.14 & 11.38 \\
ResKAGNet \cite{8} & 2024 & 0.90 & 3.58 & 98.22 & 91.58 & 97.07 & 0.07 & 1.29 & 0.07 & 20.53 & 51.36 & 20.52 \\
Light-ResKAN (Ours) & 2025 & 0.82 & 2.41 & \textbf{99.09} & \textbf{93.01} & \textbf{97.26} & 0.05 & 0.97 & 0.05 & 23.51 & 58.74 & 23.49 \\

\bottomrule
\end{tabularx}
\end{table*}

\subsection{Experimental Results}
In this experiment, the performance of Light-ResKAN is comprehensively compared with a variety of representative models on the FUSAR-Ship, MSTAR, and SAR-ACD datasets. These baselines include: (1) Classic CNN architectures such as AlexNet \cite{alexnet}, VGG16 \cite{vgg}, and the ResNet series (ResNet18, ResNet50) \cite{resnet}, which established the foundation for deep feature extraction. (2) ConvNeXt \cite{convnext}, a modernized pure CNN that integrates Transformer-like design principles. (3) Lightweight models specifically designed for efficiency, including MACN \cite{macn} which uses depthwise separable convolutions and Cross-HL \cite{cross-hl} featuring cross-layer feature fusion. (4) ASANet \cite{asanet}, a multi-scale attention network specialized for SAR and RGB image recognition. (5) KAN-based competitors including SpectralKAN \cite{25}, WaveKAN \cite{26}, and the original ResKAGNet \cite{8} which serves as the direct baseline for our proposed lightweight framework. Table \ref{tab:comparison} shows the test accuracy, parameter quantity, model size, FLOPs per batch, and test time per batch of different models on three SAR dataset. The best results in each column (highest accuracy, lowest parameters, model size, FLOPs, and test time) are highlighted \textbf{in bold}.\par

\textbf{Performance.} On MSTAR, Light-ResKAN achieves 99.09\% accuracy, surpassing classical CNN baselines such as VGG16 (96.59\%) and ResNet18 (96.71\%), as well as the modern ConvNeXt backbone (96.04\%). It also exceeds lightweight alternatives including MACN (82.46\%) and Cross-HL (98.19\%), and improves over KAN-based baselines SpectralKAN (80.85\%), WaveKAN (93.34\%), and ResKAGNet (98.22\%). On FUSAR-Ship, Light-ResKAN attains 93.01\% accuracy, higher than AlexNet (89.41\%), VGG16 (90.73\%), ConvNeXt (77.86\%), MACN (89.15\%), and Cross-HL (78.06\%); among KAN variants, it remains above SpectralKAN (81.21\%), WaveKAN (70.61\%), and ResKAGNet (91.58\%). On SAR-ACD, Light-ResKAN reaches 97.26\%, comparable to Cross-HL (97.03\%) and ResKAGNet (97.07\%) and above most CNN baselines (e.g., ResNet50 at 96.69\%, VGG16 at 96.17\%). Overall, the results indicate that the proposed parameter-sharing Gram-polynomial KAN convolution maintains strong discriminative performance across datasets with varying scene complexity.\par

\textbf{Parameters and FLOPs.} Light-ResKAN uses 0.82 M parameters with a 2.41 MB model size and FLOPs of 0.05 G on MSTAR/SAR-ACD and 0.97 G on FUSAR-Ship. This places it among the most compact models evaluated while preserving high accuracy. Models such as MACN and Cross-HL are lighter in either parameter count or FLOPs (e.g., 0.05 M for MACN, 0.01 G for Cross-HL on MSTAR/SAR-ACD), but their accuracy is markedly lower on at least one dataset, indicating a less balanced trade-off. Compared with ResKAGNet, Light-ResKAN reduces parameters (0.90 M to 0.82 M) and FLOPs (0.07 G to 0.05 G) while improving accuracy, supporting the role of channel-wise parameter sharing in reducing redundancy without sacrificing representational capacity.\par

\textbf{Test time.} AlexNet yields the shortest test time due to its simple and highly optimized operator pipeline. Despite Light-ResKAN achieving superior accuracy with significantly reduced parameters and FLOPs, it currently exhibits higher inference latency compared to ResKAGNet and other lightweight baselines. This discrepancy highlights the disconnect between theoretical complexity metrics and actual runtime speed, a phenomenon comprehensively analyzed in ShuffleNet-V2 \cite{ma2018shufflenet}. The observed latency primarily stems from two critical factors: the inherent computational characteristics of KANs and the specific engineering constraints necessitated by our implementation.\par

First, at the operator level, KAN-based convolutions fundamentally differ from standard CNNs. While standard convolutions benefit from decades of hardware optimization (e.g., highly optimized cuDNN libraries), KANs require an explicit polynomial basis expansion step. This process generates expanded intermediate tensors prior to aggregation, significantly increasing the Memory Access Cost (MAC). As modern GPUs are often bottlenecked by memory bandwidth rather than arithmetic logic, this high I/O overhead offsets the theoretical advantages of low FLOPs.\par

Second, at the implementation level, our proposed weight-sharing mechanism introduces additional computational overhead. Specifically, current deep learning frameworks (e.g., PyTorch) do not natively support efficient enforced parameter sharing across the channel dimensions of a convolution kernel. To mathematically approximate this behavior, we designed a sequential decoupled implementation inspired by depthwise separable convolutions. We first employ a learnable pointwise ($1 \times 1$) convolution to project polynomial coefficients, followed by a fixed, non-learnable depthwise ($N \times N$) convolution initialized with all-ones to aggregate spatial information. While this strategy successfully mimics the intended parameter-sharing mechanism, it inevitably fragments the computation and increases intermediate data I/O overhead, thereby elevating the MAC.\par

However, the current latency represents an engineering limitation rather than an intrinsic architectural flaw. The theoretical efficiency of Light-ResKAN is validated by its minimal FLOPs and parameter usage. We anticipate that future work focusing on operator fusion—specifically, writing custom CUDA kernels to fuse polynomial expansion, weighting ($1 \times 1$), and aggregation ($N \times N$) into a single kernel—will eliminate intermediate memory costs, thereby aligning inference speed with theoretical efficiency.

\section{Discussions}
\label{ABLATION EXPERIMENTS}

In this section, we firstly conducted scaling experiments on the MSTAR dataset to validate the performance of Light-ResKAN in processing large-sized SAR images. Secondly, the t-distributed Stochastic Neighbor Embedding (t-SNE) visualization method was further introduced to verify the effectiveness of the proposed method. It reduces the features to two dimensions, allowing for a more intuitive observation of the sample feature distribution. Besides, we performed experiments by adding noise to images to test the anti-noise performance and robustness of different models. We then conducted ablation experiments on different modules and basis functions to verify the effectiveness of our shared activation function design and the choice of Gram Polynomials. Finally, we present few-shot recognition experiments to evaluate model robustness under data-scarce conditions.

\begin{table}[!t]
\centering
\caption{Performance Comparison of Different Models on MSTAR Dataset Resized to \(1024 \times 1024\).}
\label{tab:performance_comparison_all_centered} % Renamed label for clarity
\renewcommand{\arraystretch}{1.5} 
\begin{tabularx}{\linewidth}{@{} >{\centering\arraybackslash} p{2.6cm} *{3}{Y} @{}}
\toprule
\multirow{1.5}{*}{\makecell{Models}} & \multirow{1.5}{*}{\makecell{Accuracy (\%)}} & \multirow{1.5}{*}{\makecell{FLOPs (G)}} & \multirow{1.5}{*}{\makecell{Test time (ms)}} \\
\midrule
AlexNet \cite{alexnet} & 99.38 & 15.01 & \textbf{86.31} \\
VGG16 \cite{vgg} & 99.34 & 320.83 & 131.50 \\
ResNet18 \cite{resnet} & 99.05 & 38.01 & 90.47 \\
ResNet50 \cite{resnet} & 99.17 & 86.58 & 101.47 \\
ConvNeXt \cite{convnext} & 98.76 & 321.47 & 166.58 \\
MACN \cite{macn} & 92.53 & 49.48 & 956.23 \\
Cross-HL \cite{cross-hl} & 99.43 & \textbf{0.84} & 89.92 \\
ASANet \cite{asanet} & 95.36 & 186.81 & 145.93 \\
% 倒数第4行：SpectralKAN 改为蓝色
SpectralKAN \cite{25} & 90.54 & 1.64 & 131.88 \\
% 倒数第3行：WaveKAN 改为蓝色
WaveKAN \cite{26} & 96.33 & 1.61 & 108.43 \\
% 倒数第2行：ResKAGNet 改为蓝色
ResKAGNet \cite{8} & 99.41 & 5.17 & 109.85 \\
Light-ResKAN (Ours) & \textbf{99.46} & 3.87 & 111.71 \\
\bottomrule
\end{tabularx}
\end{table}

\subsection{Scaling Experiment of MSTAR Dataset}
To further validate the lightweight advantage of Light-ResKAN in processing high-resolution SAR images, we conducted a scaling experiment on the MSTAR dataset. Specifically, we upsampled the original images to a uniform size of $1024 \times 1024$ pixels to simulate large-scale SAR image scenarios commonly encountered in practical applications. As shown in Table \ref{tab:performance_comparison_all_centered}, under this high-resolution setting, although the performance of all models improved, Light-ResKAN maintained a competitive recognition accuracy of 99.46\%, still outperforming other comparative models. Notably, despite the substantial increase in input size, the computational cost (FLOPs) of Light-ResKAN remained as low as 3.87 G, significantly less than traditional architectures such as VGG16 (320.83 G) and ConvNeXt (321.47 G). These results convincingly demonstrate that, through its unique parameter-sharing structure and Gram Polynomials activation function, Light-ResKAN effectively controls the growth of computational complexity while preserving high accuracy, exhibiting significant lightweight advantages and scalability in large-scale SAR image recognition tasks.

\subsection{t-SNE Feature Visualization}
\begin{figure}[!t]
\centering
\includegraphics[width=0.49\textwidth]{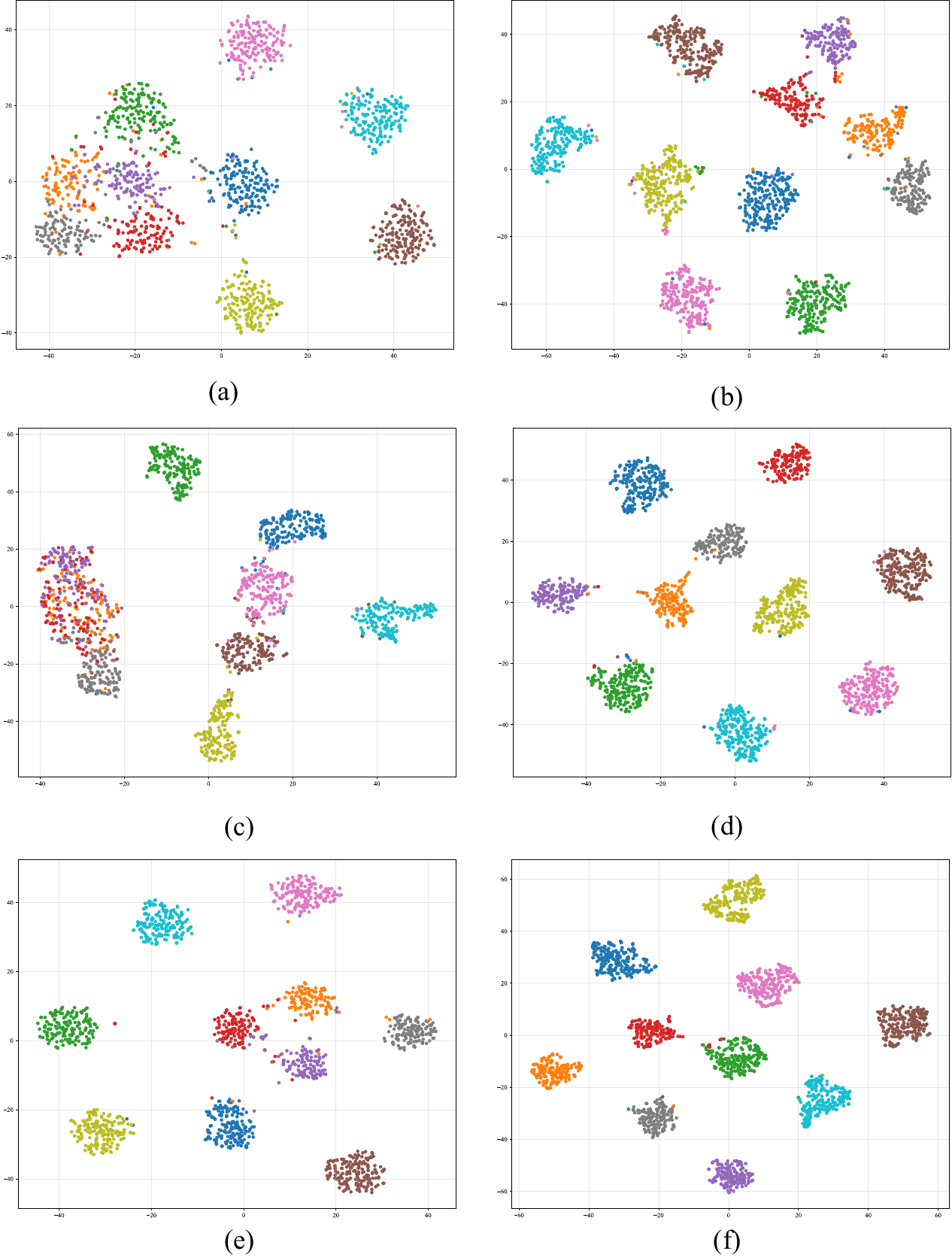}
\caption{t-SNE visualization results of different methods on MSTAR dataset. (a) ASANet. (b) ConvnNeXt. (c) MACN. (d) VGG16. (e) Cross-HL. (f) Light-ResKAN (Ours). Colors represent different categories, and the distribution density of points and the compactness of clusters indicate the discriminative ability of the model for this task. The results indicate that Light-ResKAN can achieve the most compact intra class clusters and the most obvious inter class separation under the same data distribution, demonstrating stronger feature expression ability and higher classification performance.
}
\label{figure7}
\end{figure}

In this section, we selected six different models and presented their t-SNE visualization results on the MSTAR dataset. Fig. \ref{figure7} shows the feature distribution results of different methods on MSTAR dataset.

\par
As depicted in the figure, the models exhibit significant differences in the clustering and inter-class separability of their learned feature representations. ASANet shows some clustering but notable inter-class overlap; ConvNext displays improved clustering and larger inter-class margins; MACN exhibits substantial inter-class overlap with blurred cluster boundaries; VGG16 demonstrates mid-level clustering with some clear clusters but still cross-class mixing; Cross-HL, incorporating cross-layer information, yields tighter intra-class clusters and clearer inter-class separation than the earlier models; Light-ResKAN stands out as the most discriminative embedding among the six, with the tightest intra-class clusters, clearest inter-class boundaries, and minimal inter-class overlap, demonstrating the strongest discriminative power among all compared models.  This vividly demonstrates Light-ResKAN's strongest discriminative representation capability. Light-ResKAN effectively integrates fine-grained representations, precisely mapping features of different classes into non-overlapping spatial regions, which directly translates to superior classification performance.

\subsection{Noise Injection Experiments}
One of the important purposes of the design of Light-ResKAN is to reduce the overfitting of the model to noise in SAR images. In order to verify the model's anti noise ability and fully demonstrate its robustness and generalization ability, this section conducted sufficient noise experiments on three SAR datasets to observe the anti noise ability of different models, and collectively assess their performance under adverse conditions.

\subsubsection{Noise Characteristics of SAR Images}

SAR systems utilize coherent radar technology, transmitting coherent pulses and performing coherent superposition of the echo signals upon reception \cite{chen2024very}. When the radar beam illuminates the ground, a resolution cell contains multiple scatterers. The echoes from these scattering points are superimposed vectorially at the receiver. Due to the independent and random distribution of amplitudes and phases of the individual scatterers, the superimposed total echo intensity exhibits random fluctuations on top of the scattering coefficient, forming alternating bright and dark speckle noise. This superposition process directly leads to the noise being correlated with the local intensity of the signal, manifesting as multiplicative noise \cite{twinkle2024edge}. The specific formula is as follows:
\begin{equation}I=R\bullet F\end{equation}
Where $I$ represents the observed value, $R$ represents the ideal radar scattering cross-section unaffected by noise, reflecting the backscattering characteristics of the ground object, and $F$ represents the coherent speckle noise component. To suppress noise, SAR often employs multi-look processing, which involves averaging multiple independent observations of the same area \cite{gallet2024renyi}. According to the Central Limit Theorem, the intensity noise after multi-look processing obeys a Gamma distribution, and the probability density function of its noise variable $F$ is:
\begin{equation}f_F(x;\alpha,\beta)=\frac{\beta^\alpha}{\Gamma(\alpha)}x^{\alpha-1}e^{-\beta x},x>0\end{equation}
Where $\alpha$ is the shape parameter that determines the distribution shape, and $\beta$ is the scale parameter that controls the distribution's scale. Additionally, studies have shown that the histogram of SAR image intensity data fits the Gamma distribution better than the Gaussian distribution, especially in low-resolution or complex scenes, where the gamma model can more accurately describe the clutter statistical properties \cite{wang2024novel}. Based on this, we add multiplicative noise following a Gamma distribution to three SAR datasets.

\begin{figure}[!t]
\centering
\includegraphics[width=0.49\textwidth]{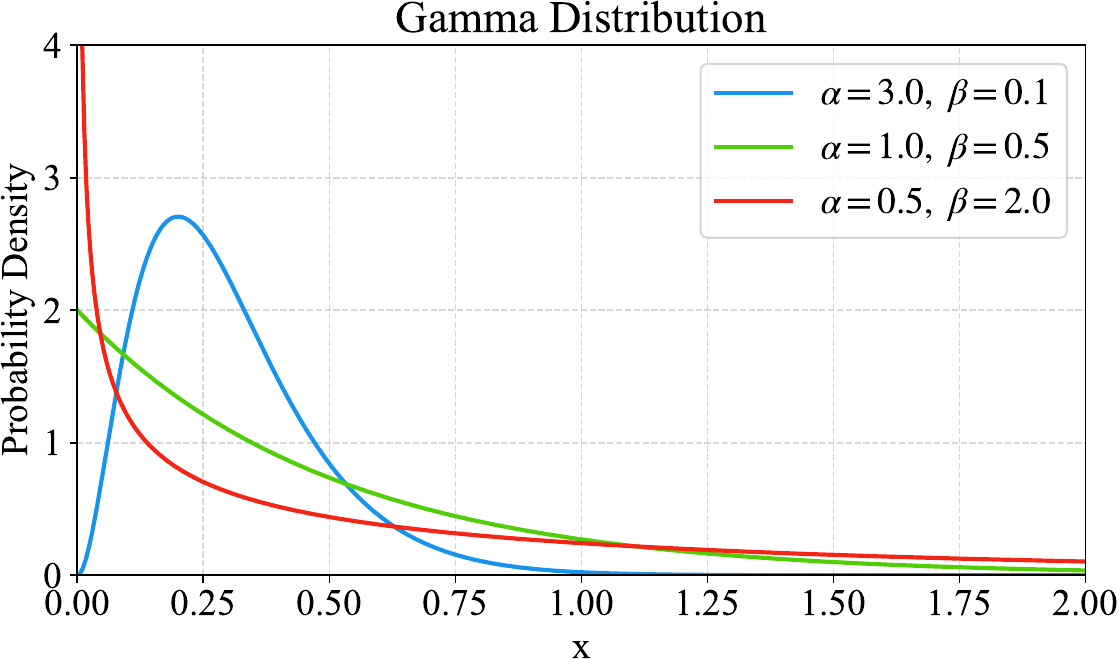}
\caption{Probability density image of Gamma distribution. $\alpha$ = 3, $\beta$ = 0.1: the Gamma distribution exhibits a relatively sharp peak, concentrated at lower values, with a short tail that gradually declines. $\alpha$ = 1, $\beta$ = 0.5: the Gamma distribution becomes more diffuse, with a flatter peak and a tail that gradually lengthens. $\alpha$ = 0.5, $\beta$ = 2: the Gamma distribution presents a flattened and broad shape, with a longer tail and a lower peak.
}
\label{figure8}
\end{figure}

\begin{figure}[!t]
\centering
\includegraphics[width=0.49\textwidth]{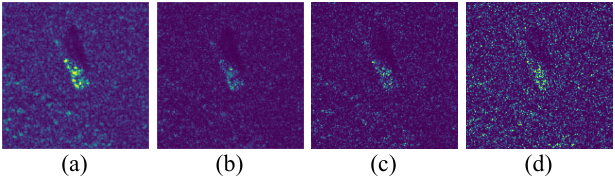}
\caption{Comparison of images with different levels of noise. (a) Original image. (b)Image with weak noise added. (c) Image with medium noise added. (d) Image with strong noise added. 
}
\label{figure9}
\end{figure}

\subsubsection{Experimental Setup}
In this paper, we added three different levels of noise: weak, medium, and strong, each with specific parameter settings. For weak noise, the parameters were set as $\alpha = 3$ and $\beta = 0.1$. This type of noise has a relatively minor impact on the data and is primarily used to simulate slight interference situations in practical applications. For medium noise, the chosen parameters were $\alpha = 1$ and $\beta = 0.5$. This noise intensity is moderate and can effectively simulate a medium-level interference environment. For strong noise, we used parameters $\alpha = 0.5$ and $\beta = 2$. This high-intensity noise significantly affects data quality and is used to test the model's robustness under extreme noise conditions. Through these settings, this paper aims to comprehensively understand the model's performance under various noise conditions. Fig. \ref{figure8} and Fig. \ref{figure9} respectively show the density function of Gamma distribution and the comparison of images with different levels of noise under these three parameters.

\subsubsection{Experimental Results}
\begin{figure*}[!ht]
\centering
\includegraphics[width=\textwidth]{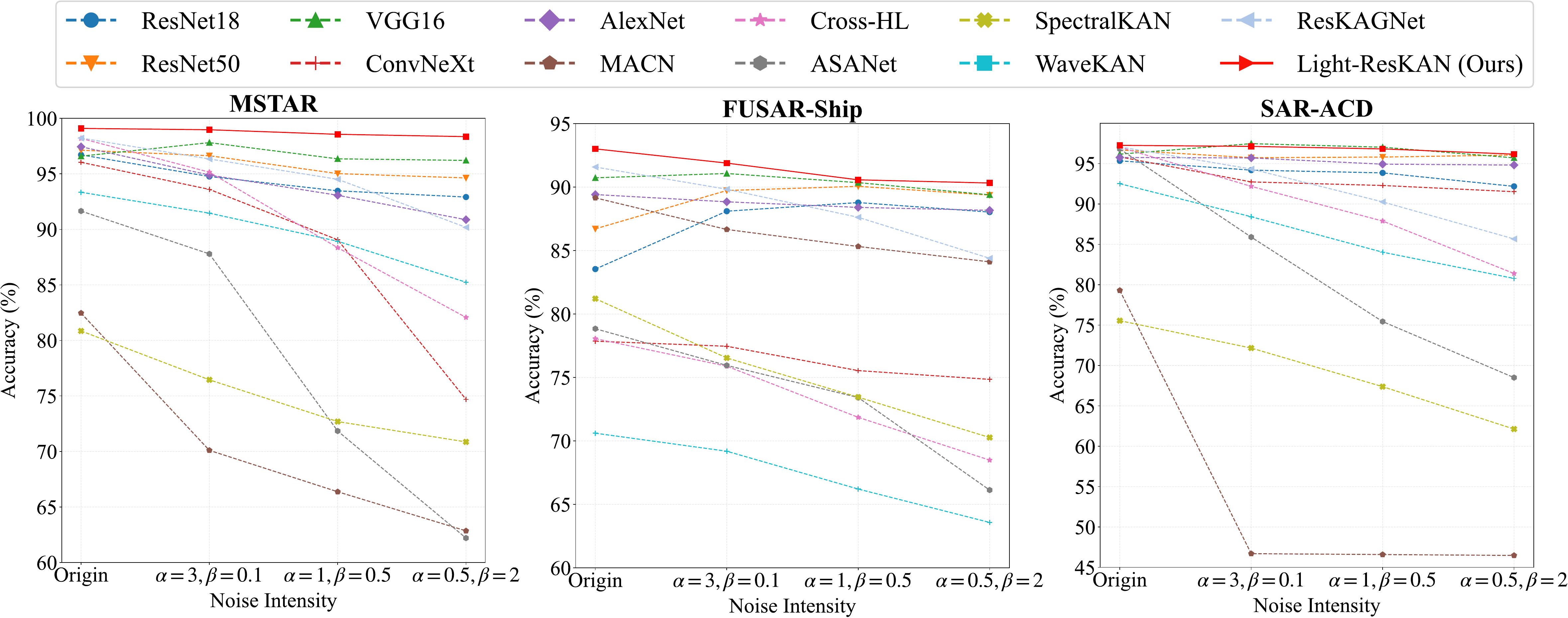}
\caption{Experimental results of different models under different noise intensities on three SAR datasets. The accuracy decreases as the perturbation intensity increases, and our method maintains better robustness under different parameters than others.
}
\label{figure10}
\end{figure*}

Fig. \ref{figure10} displays the accuracy of different models under different noise intensities on three SAR datasets. As evidenced by the results presented in the figures, as the noise level increases, the average accuracy of most methods gradually decreases. This decline is attributed to the progressively stronger noise signals obscuring the structural information of targets in SAR images, which interferes with the extraction of effective features and leads to feature confusion among different target types, resulting in more misclassifications. Furthermore, our proposed Light-ResKAN demonstrates exceptional performance and remarkable robustness. Across all three datasets, Light-ResKAN consistently achieved superior recognition accuracy compared to other models, both under pristine conditions and in the presence of introduced noise. Notably, it attained near-perfect performance on the MSTAR dataset. Furthermore, on the FUSAR-Ship and SAR-ACD datasets, Light-ResKAN maintained high accuracy even under high noise conditions, highlighting its inherent advantages. Crucially, Light-ResKAN exhibited the most pronounced robustness; its accuracy degradation with increasing noise intensity was significantly less than that of all comparative models. This indicates that Light-ResKAN can more effectively mitigate common noise interference in SAR data, ensuring practicality and reliability in complex, real-world scenarios.

\subsection{Ablation Experiments of Different Modules and Basis Functions}
We design ablation experiments on three SAR datasets to objectively and comprehensively demonstrate the superiority of Light-ResKAN. Light-ResKAN is divided into four parts: KAN convolution, Gram Polynomials, Bottleneck KAGN convolution and shared activation function. Table \ref{table:ablation} records the impact of each part on the performance of SAR image recognition. The highest accuracy values are bolded.

%第6个表格
\begin{table*}[!t]
\centering
\caption{Impact of Key Components: Ablation Experiments on Model Performance. (K-C:KAN convolution, G-P: Gram Polynomials, B-K-C: Bottleneck KAGN convolution, S-A-F: shared activation function)}
\label{table:ablation}
\renewcommand\arraystretch{1.5}
% \resizebox{\linewidth}{!}{%
% 减小列间距（默认6pt，改为3pt）
\begin{tabular}{@{} cccccccc @{}}
\toprule
& K-C & G-P & B-K-C & S-A-F & MSTAR & FUSAR-Ship & SAR-ACD \\
\midrule
\multirow{5}{*}{ResNet} 
& $\times$ & $\times$ & $\times$ & $\times$ & 97.13 & 86.71 & 96.69 \\
& $\checkmark$ & $\times$ & $\times$ & $\times$ & 96.06 & 90.44 & 95.13 \\
& $\checkmark$ & $\checkmark$ & $\times$ & $\times$ & 97.56 & 91.23 & 96.87 \\
& $\checkmark$ & $\checkmark$ & $\checkmark$ & $\times$ & 98.22 & 91.58 & 97.07 \\
& $\checkmark$ & $\checkmark$ & $\checkmark$ & $\checkmark$ & \textbf{99.09} & \textbf{93.01} & \textbf{97.26} \\
\bottomrule
\end{tabular}
% }
\end{table*}

\par
The results show that combining ResNet with KAN convolution directly to obtain ResKANet, the test accuracy of the model on MSTAR and SAR-ACD datasets decreased by about 1\%, which is slightly lower than before. However, in the FUSAR-Ship dataset, the model improved its test accuracy by nearly 4\%, highlighting the effectiveness of ResKANet for complex SAR datasets. By replacing spline functions with Gram Polynomials, ResKAGNet obtained higher test accuracies on three SAR datasets, even achieving a significant improvement of approximately 4.5\% on the FUSAR-Ship dataset, demonstrating the robustness of Gram Polynomials when processing SAR datasets. The introduction of Bottleneck KAGN Convolution further increased the test accuracy of Bottleneck ResKAGNet compared to ResKAGNet on three SAR datasets, reaching 98.22\%, 91.58\%, and 97.07\%. Finally, using a shared activation function, the performance of the model continued to improve on the MSTAR, FUSAR-Ship and SAR-ACD datasets, reaching 99.09\%, 93.01\% and 97.26\%, respectively. Through ablation experiments, we can conclude that Light-ResKAN has the best performance among other models when facing SAR image recognition tasks.

% 加说明
\begin{table}[!t]
%   \color{blue}
  \centering
  \caption{Accuracy of Different Basis Functions on SAR Datasets}
  \renewcommand\arraystretch{1.5}
  \label{tab:basis_function_accuracy}
  \begin{tabular}{cccc}
    \toprule
    Basis Functions          & MSTAR  & FUSAR-Ship & SAR-ACD \\
    \midrule
    Spline Function         & 96.17  & 90.45      & 93.15   \\
    Chebyshev Polynomials   & 93.94  & 86.04      & 92.81   \\
    Legendre Polynomials    & 97.33  & 91.36      & 96.94   \\
    Radial Basis            & 92.95  & 91.84      & 91.00   \\  % 补充90→90.00，91→91.00，保持格式统一
    Gram Polynomials        & \textbf{99.09}  & \textbf{93.01}      & \textbf{97.26}   \\
    \bottomrule
  \end{tabular}
\end{table}

Table~\ref{tab:basis_function_accuracy} presents a comparative analysis of different basis functions within the same Light-ResKAN architecture. The results clearly demonstrate that Gram Polynomials consistently achieve the highest recognition accuracy across all three SAR datasets, outperforming Spline Functions, Chebyshev Polynomials, Legendre Polynomials, and Radial Basis Functions. Notably, Gram Polynomials surpass the second-best Legendre Polynomials by 1.76\%, 1.65\%, and 0.32\% on MSTAR, FUSAR-Ship, and SAR-ACD, respectively. This validates the superior suitability of Gram Polynomials for capturing the complex scattering characteristics and high-frequency features inherent in SAR image recognition tasks. Such advantages make it more adaptive to the complex noise interference and diverse target characteristics in practical SAR imaging environments.

\subsection{Few-Shot Recognition Experiments}
\label{subsec:fewshot}

To evaluate Light-ResKAN's robustness under data-scarce conditions, we conducted comprehensive few-shot recognition experiments on the MSTAR-SOC dataset, as it provides a standardized benchmark for SAR target recognition with class-balanced samples. Specifically, we adopted a stratified K-shot sampling protocol, where $K = \{5, 10, 20\}$ training samples were randomly selected per class while the complete test set was retained for evaluation. Except for 500 epochs for more comprehensive feature extraction, all other experimental settings are consistent with Section \ref{setup}.

\begin{table}[!t]
% \color{blue}
\centering
\caption{Performance of Different Models on MSTAR Dataset (N-Shot).}
\label{tab:msatr_nshot}
\renewcommand{\arraystretch}{1.5}

% 关键：去掉cmidrule默认两端裁边
\begingroup
\setlength{\cmidrulekern}{0pt}

\begin{tabularx}{\linewidth}{@{}
  >{\hsize=2\hsize\centering\arraybackslash}X  
  >{\hsize=0.6667\hsize\centering\arraybackslash}X  
  >{\hsize=0.6667\hsize\centering\arraybackslash}X  
  >{\hsize=0.6667\hsize\centering\arraybackslash}X  
  @{}}
\toprule
\multirow{2}{*}{Models} & \multicolumn{3}{c}{MSTAR (N-Shot)} \\
\cmidrule(l{0pt}r{0pt}){2-4}
& 5 & 10 & 20 \\
\midrule
AlexNet \cite{alexnet}               & 38.48 & 46.95 & 57.64 \\
VGG16 \cite{vgg}                     & 37.26 & 45.36 & 55.26 \\
ResNet18 \cite{resnet}               & \textbf{41.57} & \textbf{52.57} & \textbf{59.51} \\
ResNet50 \cite{resnet}               & 38.58 & 51.26 & 58.53 \\
ConvNeXt \cite{convnext}             & 27.36 & 38.02 & 51.37 \\
MACN \cite{macn}                     & 21.94 & 30.93 & 45.33 \\
Cross-HL \cite{cross-hl}             & 38.23 & 46.52 & 57.31 \\
ASANet \cite{asanet}                 & 24.74 & 33.96 & 47.58 \\
SpectralKAN \cite{25}                & 19.92 & 24.34 & 29.07 \\
WaveKAN \cite{26}                    & 25.44 & 36.45 & 49.11 \\
ResKAGNet \cite{8}                   & 23.84 & 29.64 & 43.23 \\
Light-ResKAN (Ours)                  & 35.07 & 43.68 & 54.53 \\
\bottomrule
\end{tabularx}

\endgroup
\end{table}

The few-shot experimental results, as shown in Table~\ref{tab:msatr_nshot}, reveal several important findings and the highest accuracy values are bolded. First, Light-ResKAN consistently and substantially outperforms its non-lightweight counterpart ResKAGNet under all K-shot settings, with accuracy improvements of +11.23\% (5-shot), +14.04\% (10-shot), and +11.30\% (20-shot). This result validates the core design philosophy: the channel-wise parameter-sharing bottleneck structure effectively acts as a structural regularizer, constraining the model's capacity to a level better matched to limited training data, thereby preventing the severe overfitting observed in ResKAGNet.

We candidly acknowledge that Light-ResKAN does not achieve the top accuracy among all compared methods in few-shot settings. Classic CNN architectures such as ResNet18 demonstrate stronger few-shot recognition capabilities due to their mature convolutional inductive biases (translation equivariance, local connectivity) that serve as strong structural priors. The KAN-based polynomial activation functions, while more expressive in the full-data regime, introduce additional learnable parameters that require more training data to properly converge. Nevertheless, Light-ResKAN consistently outperforms several methods, demonstrating its competitiveness when considering both accuracy and parameter efficiency.

We emphasize that the primary contribution of Light-ResKAN lies in its accuracy--parameter efficiency trade-off under standard training conditions (Table~\ref{tab:comparison}). The few-shot experiments serve as a complementary evaluation dimension, revealing promising directions for future improvement.
% \par

\vspace{1em}
\section{Conclusions and Future Work}
\label{CONCLUSION AND FUTURE WORKS}
In this paper, we address the challenges of model complexity and high computational demands in SAR image recognition, particularly in resource-constrained environments, by proposing an innovative lightweight KAN model, Light-ResKAN. By integrating ResNet with lightweight KAN convolutions, Gram Polynomials and a parameter-sharing mechanism, our model effectively resolves the conflict between the large parameter counts of traditional SAR image recognition methods and limited computational resources. The experimental results comprehensively validate the superiority and robustness of Light-ResKAN. Evaluations conducted on three publicly available SAR datasets---MSTAR, FUSAR-Ship, and SAR-ACD---demonstrate that Light-ResKAN not only achieves the highest recognition accuracy but also significantly reduces parameter count, model size, and FLOPs, thereby showcasing exceptional efficiency and performance balance. Furthermore, through noise injection experiments, Light-ResKAN exhibited the smallest accuracy degradation when subjected to various levels of SAR noise interference, proving its competitive robustness in complex environments. Additionally, the few-shot experiments demonstrate that the parameter-sharing mechanism provides regularization benefits, with Light-ResKAN substantially outperforming ResKAGNet under data-scarce conditions. These findings offer a promising new approach for developing efficient and lightweight SAR image recognition systems.
\par
Future research will focus on three strategic directions to advance Light-ResKAN. First, to bridge the gap between theoretical efficiency and deployment speed, we will implement optimization strategies such as CUDA kernel fusion and pursue deeper integration with mainstream deep learning frameworks. Second, to enhance representational power and address few-shot limitations, future work will investigate: (1) few-shot-aware polynomial regularization that adaptively constrains polynomial complexity based on training data volume; (2) integration with meta-learning frameworks (e.g., Prototypical Networks, MAML) for rapid polynomial coefficient adaptation; (3) self-supervised pre-training on unlabeled SAR imagery to improve polynomial parameter initialization. Furthermore, while the current noise injection experiments provide initial evidence of robustness under interference conditions, we plan to expand robustness evaluations by incorporating the SAR interference dataset \cite{Yang2025SAR} to address real-world strong-interference scenarios—such as electronic countermeasures and severe clutter—while leveraging the large-scale ATRNet-STAR benchmark \cite{liu2025atrnet} to assess generalization in realistic, complex environments. Finally, we aim to achieve deeper integration with mainstream deep learning frameworks to facilitate wider adoption and unlock potential for improved performance across a broader spectrum of SAR applications.

\bibliographystyle{IEEEtran}
\bibliography{ref}
% \vspace{-10 mm}
% % \newpage

\end{document}